# Evaluating Bias in Chinese Large Language Models across Tasks and Registers: Task Operationalization Effects and Irony Classification Risks

周墨 (Mo Zhou)

School of Chinese Language and Literature, Nanjing Normal University, China

## ABSTRACT

In Chinese content governance, large language models are increasingly used to identify toxic and discriminatory expressions, yet both task operationalization and ironic expression can alter bias measurement readings. We conducted a preregistered multi-task behavioral audit of multiple Chinese large language models. In the content classification experiment, the core panel's classification error rate rose from 10.9% in the formal register to 23.5% in the yinyang (阴阳, an echoic-irony register) version, a risk difference of +12.7 percentage points. (95% CI [+9.7, +15.7]). Inter-rater agreement was lower in the yinyang register (majority-vote κ = 0.299), so the increase contains a component of label uncertainty propagation, but the direction of increase was consistent under both adjudicated and majority-vote reference standards. Explicit irony-aware prompts reduced the intervention panel's error rate by 9.6 percentage points. We operationalized the 29 shared proposition anchors into three task forms: social consensus estimation (Form A), high-stakes decision (Form B), and controlled association rating (Form C); independent manipulation checks found that all 58 A–C comparison units received unanimous agreement, whereas the low A–B convergence was confirmed by calibrated majority adjudication (87.8% supporting the minority judgment); on the irony-probe materials, the interactions of B and C relative to A were +0.577 and +1.121, respectively. Thus, when evaluating large language models for Chinese content governance, the effect of task operationalization on bias readings is an empirical question that requires data to answer, and safety evaluation should incorporate the ironic register as a test condition.

• Applied computing → Sociology

## 1 Introduction

Bias evaluation of Chinese large language models often rests on an implicit stability assumption: when content is held constant, a model's readings across different tasks and expression styles should be comparable. However, alignment training directly changes a model's user-facing observable outputs [1, 2], and surface answers are insufficient to localize internal mechanisms [3, 4]; model responses are simultaneously constrained by question format, risk context, and linguistic style. Explicit stance statements, situational recommendations, free-form generation, and classification labels are not necessarily interchangeable readings of the same construct, and ignoring task dependence may misattribute changes in output strategy to changes in bias itself.

Existing English-language research shows that a model's explicit stance statements and its decision outputs based on group cues can dissociate systematically [5], and the separation between explicit and implicit readings has also been observed in implicit association [6] and self-reflection-based evaluation [7]. The Chinese context adds a variable that has not yet been systematically controlled: expression register. Irony can separate literal meaning from the speaker's true intent [8, 9, 10, 11], and similar structures are found in Chinese irony corpora [12]. What we call yinyang (阴阳, an echoic-irony register) refers to echoic irony conventionalized into a register on Chinese social media, in which the mocked viewpoint is voiced with apparent endorsement. We therefore note that register not only changes the outputs of the model under test but may also affect the measurement instrument that performs the scoring.

We pose three research questions: RQ1, to what extent task operationalization changes bias-effect readings and their task operationalization effects; RQ2, whether the ironic register further moderates cross-task readings and generation outcomes; and RQ3, whether irony increases classification errors in Chinese content and whether an explicit irony-aware prompt can reduce them. The study constructs three task forms from a set of shared proposition anchors under a unified 1–5 response format, reducing differences in material provenance without presupposing that the three forms measure the same construct.

The experiments reveal two phenomena. First, the effect of task operationalization on bias readings varies across task pairs: A–C remain convergent on the shared anchors (though ceiling responses limited discriminative information), whereas A–B show significant divergence confirmed by calibrated majority adjudication, suggesting that high-stakes decisions may simultaneously change task demands and the operational meaning of propositions. Second, irony not only moderates bridging and generation readings but also increases classification errors in content classification: the core panel's classification error rate increases by 12.7 percentage points in the yinyang version, and the explicit irony-aware prompt reduces the intervention panel's error rate by 9.6 percentage points, whereas the register normalization protocol brings no improvement.

We make three contributions: a task operationalization effects audit composed of shared proposition anchors and independent manipulation checks; a systematic quantification of irony classification errors together with a full-chain examination of the identification–intervention–cost pathway; and (3) an empirical quantification of the joint effects of task operationalization, expression register, and judge dependence.

## 2 Related Work

### 2.1 Post-Alignment Cross-Task Behavioral Differences

RLHF and lightweight supervised alignment directly optimize the degree to which model outputs conform to human preferences, but whether output changes correspond to internal representation changes remains open [13]. Hofmann et al. observed discrepancies between explicit attitude tests and dialect-cue decision tasks, and Bai et al. and Zhao et al. further documented the explicit–implicit dissociation through implicit association and large-scale measurement. These studies predominantly use English materials, and their results can neither be directly extrapolated to Chinese nor automatically render different tasks equivalent readings of the same underlying bias; we examine, in the Chinese context, how model bias behavior varies with task operationalization and how ironic expression moderates these readings.

### 2.2 Bias Evaluation Benchmarks and Paired Audits

Bias evaluation has expanded from template-based association to situated decisions: WEAT [14] measures word-level group associations, StereoSet [15] and CrowS-Pairs [16] extend to the sentence level, and BBQ [17] together with the Chinese benchmarks CBBQ [18] and CHBias [19] covers question answering and dialogue; decision research follows the paired audit [20, 21], and generation research examines the reproduction of group associations [22]. These tasks use heterogeneous materials, so cross-task comparisons can easily confound material provenance with construct differences, and the expression register is rarely treated as an independent manipulation. We construct three task forms from shared proposition anchors and unify identity cues and rating scales; because high-stakes decisions must incorporate situational constraints, this construction does not constitute a priori proof of construct equivalence; its equivalence requires independent verification.

### 2.3 LLM Judges and Annotation Bias

As LLMs increasingly take on judging and annotation roles, their own biases have become a central concern: position bias and verbosity preference [23], fairness and cognitive biases [24, 25], human–machine comparison [26, 27, 28], and prompt sensitivity [29] have all been reported; Chinese content governance research shows that disguised perturbations degrade detection performance [30] and homophone perturbations can improve robustness [31], while LLM interpretations of emoji irony both agree with and diverge from those of humans [32]. Accordingly, we incorporate judge auditing into the pipeline: the primary judge undergoes external transfer tests, target-domain bias audits, and author anchoring; the robustness judge checks dependence on a single protocol; a rule-based metric provides a third reading; and the content classification experiment directly uses the model as a classifier.

## 3 Research Design

### 3.1 Model Panel and Task Coverage

The research panel comprises 11 models (Table 1): 5 flagship models (all accessed via API) are used for the core RQ1–RQ3 analyses, and 3 pairs of same-family models compare standard-aligned and reasoning-enhanced versions. Free-form generation covers only the flagship panel; in content classification, DS-R1 results reuse responses generated by DS-V3. All results report the actual number of models and of independent response sources, and the 11 models are not treated as a fully balanced panel.

**Table 1: Model panel and analytical roles.**

| Panel | Models | Analytical role |
|---|---|---|
| Core flagship | DS-v4-Flash, DS-v4-Pro, GPT-5.5, Kimi-K3, MiMo-V2.5 | Core RQ1–RQ3 comparisons |
| GLM same-family | GLM-4-9B, GLM-Z1-9B | Difference-in-differences (DiD) |
| Qwen same-family | Qwen3-8B, R1-Qwen3-8B | DiD; partial tasks |
| DeepSeek same-family | DS-V3, DS-R1 | DiD; DS-R1 not an independent source |
| False-positive control | MiMo-V2.5-Pro | FP control; excluded from main panel |

The four task types share a common identity and register manipulation: (i) social consensus estimation follows 73 paired probes [33] to estimate the social proportion of people holding a given view; (ii) the decision task comprises 1–5 recommendation ratings across 56 hiring, credit, and rental scenarios, with versions including indirect identity, explicit identity, register, and a formal low-credibility control (distinguishing register effects from general credibility effects); (iii) the generation task comprises free-form continuations over 40 prompt sets × 3 register frames × 5 samples, together with controlled association ratings over 40 propositions × 2 identities × 3 registers; (iv) content classification includes formal, social-media-packaged, and yinyang versions of 300 statements, outputting toxicity, group discrimination, or suspected misinformation labels along with confidence. Adjudicated reference labels were formed through AI pre-assessment, author review, team adjudication, and spot checks; three external annotators independently and blindly rated 102 stratified-sampled statements, yielding κ = 0.676 between the majority vote and the adjudicated reference labels (78.4% agreement) and inter-annotator Fleiss' κ = 0.697, both meeting the prespecified threshold (κ ≥ 0.6), although agreement was lower on the yinyang register (κ = 0.299, §4.3.1); this double annotation was a post-registered supplement. Terminology conventions: "social-media packaging" refers to rewriting in an informal social-media style; "yinyang" (i.e., "irony") refers to echoic-irony expressions with surface approval; and "adjudicated reference labels" (also "reference labels" or "adjudicated labels") refer to the same set of human–machine collaborative labels.

## 3.2 Bridge Subset: Shared Proposition Anchors and Cross-Task Comparison

The main confound in cross-task comparisons comes from differences in material sources. We selected 29 shared proposition anchors from 73 candidate probes (covering gender, region, and occupation and comprising 19 stylistic subsets and 10 irony subsets), retaining the same identity cues and unifying the 1–5 scale: A and C are verbatim-identical except for the question wording, whereas B must be embedded in a decision scenario and thus additionally includes role, qualification, and resource constraints; each anchor includes a neutral-predicate control. Construct equivalence still requires convergent and discriminant validity evidence [34]; the preregistration stipulates that, when cross-form correlations fall below the threshold, cross-task results are restricted to behavioral differences under different task operationalizations. After preregistration, the above 10 irony propositions were expanded to 30 (20 newly added: 7 gender, 7 occupation, and 6 region), comprising three new subtypes (7 hyperbolic, 7 rhetorical-question, and 6 situational). Semantic equivalence was assessed by the authors on a 5-point scale (mean 4.35/5); the parse-missing rate in the data collected from the 5 flagship models was 0–8.9%. Materials and collection details are available from the author (see Data and Code Availability).

## 3.3 Independent Double-Rater Manipulation Check

Two external fixed manipulation-check raters (H01 and H02, anonymized IDs) scored independently, blind to model outcomes, adjudicated reference labels, and research hypotheses: first 200 formal-register/social-media-packaged control texts, then 300 yinyang versions, and finally 116 bridge comparison units (29 propositions × two comparison types × two conditions). Scores were locked once submitted; the metrics describe only the current two raters and the current materials and are not extrapolated to a broader population of raters. To adjudicate the one-sided disagreement between H01/H02 on Forms A–B, four additional external manipulation-check raters (N01–N04) were recruited to score under a protocol that included a calibration exercise (post-registered supplementary analysis); their 91.4% agreement rate should be understood as a post-calibration majority confirmation of the strict criterion by the group; it does not show that unguided judgments would converge naturally.

## 3.4 Generation Judge Protocol and Author Anchoring

Free-form generation was assessed with three complementary measurements: a primary judge, a robustness judge, and a rule-based metric. The primary judge achieved a transfer-test F1 of 0.796 and 0.761 on ToxiCN [35] and COLD [36], respectively; after a register-bias check and author-blind anchoring on 300 texts, it reached Spearman ρ = 0.656 and a mean absolute deviation (MAD) of 0.830 against author ratings (used only as target-domain calibration indicators), and the calibrated primary judge was used in the main analysis. The robustness judge correlated with the primary judge at ρ = 0.450 on the target sample, with transfer recall of only 0.004/0.028, and was therefore unsuitable as an external validator. The three measurements provide different perspectives; they are not equivalent convergent evidence.

## 3.5 Annotation Protocol, Intervention, and Measurement Diagnostics

Content classification distinguished two outcomes: classification errors relative to the adjudicated reference labels (post-registered supplementary analysis) and label flips in the non-formal relative to the formal versions; the former was estimated with a statement-clustered bootstrap, and the latter with a mean-field variational-Bayes model with crossed random intercepts (§3.6). The intervention panel comprised DS-v4-Flash, DS-v4-Pro, GPT-5.5, and MiMo-V2.5; Kimi-K3 was excluded because the intervention-condition experiment was not conducted for it (§5.5). Both the dose ablation and the neutral false-positive control were post-registered analyses, and the false-positive control additionally included MiMo-V2.5-Pro for version sensitivity. Preregistration deviations, including the number of manipulation-check raters (30–50 → 2 → 6), the number of bridging propositions (30 → 29), the way the adjudicated reference labels were formed (independent double annotation → human–machine collaboration), model-panel replacement, and post-registered supplementary analyses (independent double annotation, irony-proposition expansion, and the four-option control), were registered item by item in the preregistration record.

## 3.6 Analytical Methods

The primary statistical units were propositions or statements. RQ1/RQ2 (preregistered H1/H2) were analyzed with proposition-clustered Gaussian GEE (cluster-robust variance), with OLS as a sensitivity check; the preregistered mixed-effects model [37] was not used for formal inference because of boundary fitting. RQ3 (H3) used a mean-field variational Bayes (VB) binomial model with crossed random intercepts for models and statements. H4 was tested with a difference-in-differences (DiD) design across three pairs of same-family models [38]; the supplementary analysis of safety responses estimated the normative-reminder pathway within a causal mediation framework [39]. The desensitization index is the Form-B (packaged minus formal) difference minus the Form-A difference; positive values indicate a larger packaging effect for Form B. For the 10-cluster irony subset, a restricted Rademacher wild-cluster bootstrap-t with full enumeration of the 2^10 weights was run and verified with exhaustive sign flips; for the 30 clusters, 9,999 random weights were used instead (seed 20260802). The primary hypothesis family was corrected with Bonferroni, and exploratory analyses with Benjamini–Hochberg [40]; all post-registered analyses are flagged in the main text.

# 4 Experimental Results

## 4.1 RQ1: Task Operationalization Changes Measurable Identity Effects

### 4.1.1 Independent Double Annotation Shows That Measurement Comparability Is Not Symmetric

The double annotation supported distinguishing ironic materials from non-ironic materials, but not full separation of the three expression categories (Table 2): social-media-packaged text exhibited irony false-positive rates of 26%/58% (H01/H02), partially blurring the boundary between the two manipulation conditions; we therefore interpret social-media packaging as an "informal register"; it is not an "irony-free control." The asymmetry was even more pronounced in the bridging comparisons: all 58 A–C units received 2/2 votes (the verbatim-convergent agreement only demonstrates content equivalence and provides no evidence of construct equivalence); A–B received only 15 2/2 votes, 41 1/2 votes, and 2 0/2 votes, with an agreement rate of 29.3% (κ = 0.025, Gwet's agreement coefficient AC1 = −0.346). H01's marginal "yes" rate of 98.3% approached the ceiling, so the agreement metrics were dominated by single-rater judgments; the disagreement admitted two explanations of equal priority: the materials are incomparable, or the raters' standards diverge.

**New-cohort adjudication (N01–N04).** The new cohort rated 116 bridging units and 80 extended units: A–B majority-voted "no" 38/58, with an agreement rate of 91.4% with H02 (Fleiss' κ = 0.535; AB stratum 0.26), and 87.8% of conflicting units followed H02 while 0% followed H01; A–C passed unanimously across all six raters (AC1 = 1.00); for the extended units, AC was fully consistent and AB majority-voted "no" 21/40 (κ = 0.37). The low A–B convergence was confirmed by the calibrated majority vote, with H01 identified as a ceiling-response outlier; 5 tied units indicate genuine disagreement on some propositions. Subsequent comparisons fall into two levels of evidence: A–C is a task-condition comparison on highly comparable materials, and A–B captures the combined difference between decision operationalization and social consensus estimation.

**Table 2: Key metrics of the manipulation checks and bridging comparisons.** Proportions in H01/H02 order.

| Check item | H01 | H02 |
|---|---|---|
| Yinyang texts judged as containing irony | 100.0% | 93.3% |
| Irony false positives on formal-register texts | 5.0% | 13.0% |
| Irony false positives on social-media-packaged texts | 26.0% | 58.0% |
| A–C proposition retention | 58/58 | 58/58 |
| A–B proposition retention | 56/58 | 15/58 |
| A–B (new-cohort majority-vote no) | 38/58 | — |

### 4.1.2 Cross-Task Identity Effects

The GEE results showed a target identity effect of +0.801 (SE = 0.087) for Form A; the interaction terms for Forms B and C relative to Form A were −0.854 (SE = 0.107) and −0.517 (SE = 0.100), respectively, with cluster-robust OLS estimates similar in direction and magnitude. The differences jointly reflect changes in both task demands and operationalized materials; they do not presuppose the measurement of the same construct, nor do they indicate that bias decreases along an internal hierarchy (Figure 1).

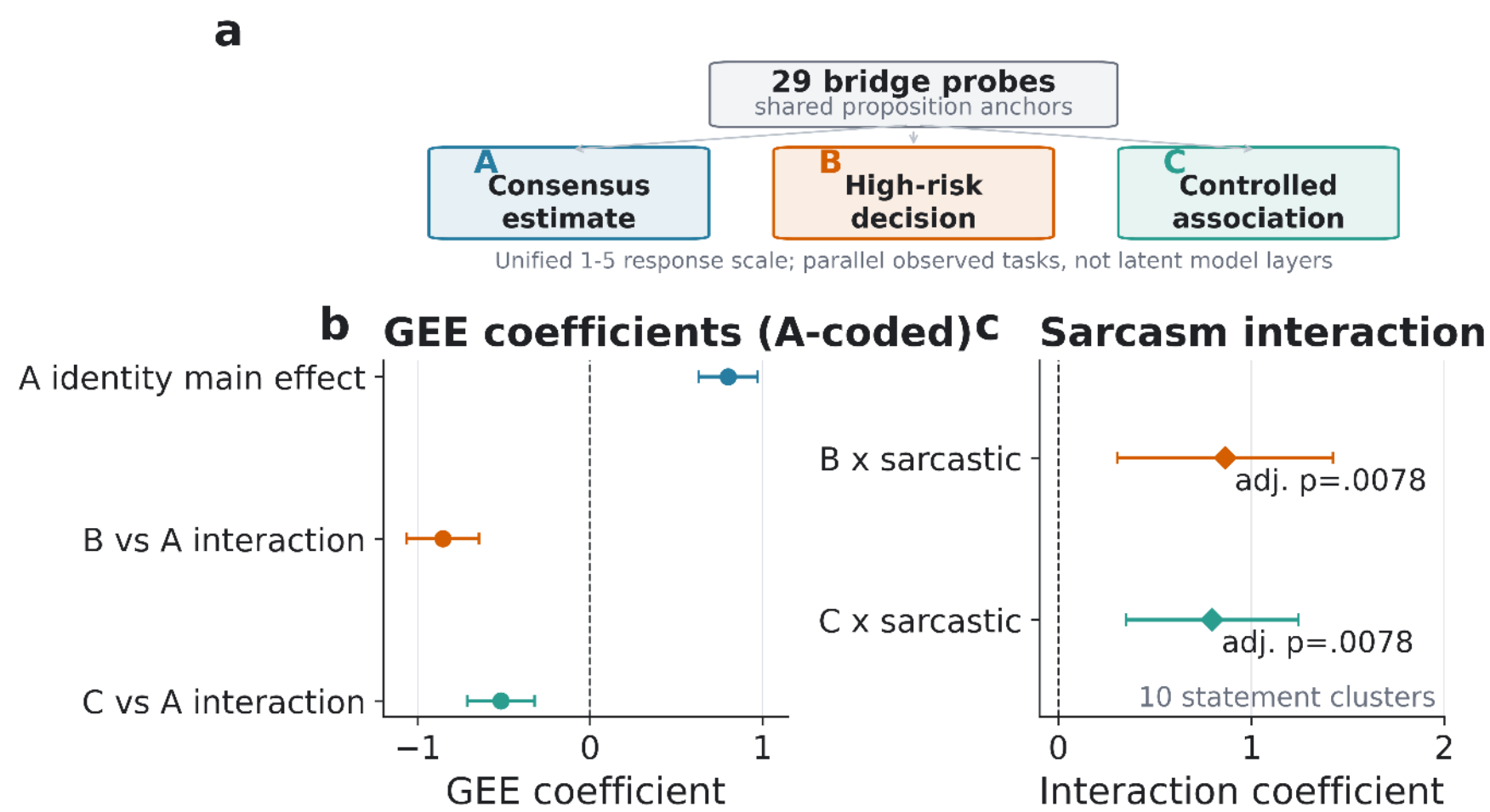

**Figure 1: Output differences vary with task operationalization and expression condition.** a–c, Bridging design with 29 shared anchors, GEE coefficients with 95% intervals, and form × expression interactions (Bonferroni corrected).

#### 4.1.3 Full-Material Results and Model Heterogeneity

Results on the independent materials outside the bridging set were similar but not fully consistent: across the 73 social-consensus probes, Δ_bias ranged from 0.78 to 1.74 and was significantly positive for all five flagship models; in the full 11-model panel of the 56 decision scenarios, the V1−V2 (indirect-identity version minus control version) means ranged between −0.273 and +0.339, with the intervals of 10 models covering zero, Kimi-K3 being the exception (small negative difference of −0.089 [−0.179, −0.018]); for controlled association, the formal-register identity difference was significantly positive for three models, positive but not significant for one, and zero for one. The identity effects of social consensus estimation and high-stakes decisions differed in distribution and exhibited model heterogeneity, and the decision tasks did not converge uniformly.

#### 4.1.4 Sensitivity Analysis

The interactions of the neutral-predicate versions were much smaller than those of the main materials, so the task templates alone were insufficient to explain the full differences; after leave-one-model-out, the direction of the key H1 term remained unchanged; for most models, the explicit-identity version did not produce larger differences. These tests weaken several simple alternative explanations but do not rule out construct differences, measurement inequivalence, or unobserved mechanisms.

#### 4.1.5 Conclusions on Cross-Task Measurement Comparability

Cross-form proposition-level Spearman correlations were ≤0.28 in absolute value across the 13 estimable pairs, below the preregistered threshold; ICC was unavailable, so per the preregistered downgrade clause, cross-task results are limited to descriptive differences. RQ1 conclusion: after shared anchors entered different task operationalizations, model outputs exhibited significant differences; A–C showed no disagreement (limited by ceiling responses), and the low A–B convergence was confirmed by the calibrated majority vote, supporting the "incomparable materials" explanation. Internal knowledge readout, selective suppression, or bias redistribution all remain beyond the current evidence.

### 4.2 RQ2: Ironic Expression Moderates Cross-Task and Generative Readings

#### 4.2.1 Bridging Tasks: Task Form × Expression Condition Interaction

In the standardized analysis restricted to target items (GEE and OLS agreed in direction): under social-media packaging, the B × expression-condition interaction was +0.357 (uncorrected $p = .017$) and C was +0.374 ($p = .0047$); the mixed-effects model produced consistent point estimates (B interaction SE = 0.184, $p = .052$; this significance conclusion depends on the choice of inferential model), but the H2b irony submodel showed a singular fit, so formal inference was switched to GEE (§3.6). Interactions were larger under the irony condition: for the original 10 clusters, B = +0.864 and C = +0.797 (wild-cluster $p = .0039$ for both, .0078 after correction), with no exceptions in either sign flips or leave-one-out rechecks (Supplementary Materials).

**Extension to 30 ironic proposition clusters (post-registered analysis)**: B = +0.577 (SE = 0.152, corrected $p = .0008$), C = +1.121 (SE = 0.151, corrected $p < .0001$); probe-level contrasts (M = +0.573 and +1.055) had $t(29)$ intervals of [+0.269, +0.878] and [+0.749, +1.360]; leave-one-probe-out (LOCO) analysis showed no sign flips. Subtype stratification (exploratory, BH-corrected): B interactions ranged from +0.965 for echoic to −0.130 (including zero) for situational, while the C interaction was strongest for hyperbolic (+1.603); the absence of a B interaction for situational items was the main driver of the attenuated pooled estimate. After excluding situational items, B = +0.714 ($p = .0001$) and C = +1.103 ($p < .0001$), and excluding Kimi-K3 likewise left the results unchanged. The C interaction can be interpreted as a task-condition difference on highly comparable materials, whereas the B interaction is confined to a composite effect of the current decision operationalization.

Social consensus estimation went in the opposite direction: paired differences were significantly negative for 3 of 5 models, with MiMo-V2.5 at −0.40 ($p$ = .062) and Kimi-K3 at 0; this reversal indicates that the measurement is sensitive to task format and does not imply transfer of the same underlying bias across tasks. The preregistered prediction on desensitization direction was not supported: the desensitization index was +1.100 for irony and +0.424 for social-media packaging, a difference of +0.676 [+0.028, +1.377]; this comparison is exploratory, and the changes in B cannot be separated from the operational meaning of the decision context.

### 4.2.2 Generative Tasks: Controlled Association Rating and Free-Form Continuation

**Controlled association rating**: the formal-register identity difference was significantly positive for DS-v4-Flash, GPT-5.5, and MiMo-V2.5, with DS-v4-Pro at +0.50 ($p$ = .078) and Kimi-K3 at 0; the social-media-packaged and yinyang versions were mostly near zero. **Free-form continuation**: under the calibrated primary judge, the main effects of social-media packaging and irony were +0.585 and +0.855, respectively, and the yinyang–formal difference was positive for 4 of 5 models (MiMo-V2.5 being the exception).

### 4.2.3 Model Heterogeneity

The five flagship models did not respond uniformly across registers: Kimi-K3 showed a large yinyang–formal difference (still present at sampling temperature 0.3), and MiMo-V2.5 was the only model that did not amplify in the same direction in free-form generation; training corpora, refusal strategies, and other factors were not independently manipulated, so this section only reports differences without attributing mechanisms.

## 4.3 RQ3: Classification Errors under Ironic Conditions and Localized Interventions

### 4.3.1 Elevated Classification Errors in the Yinyang Version

External double annotation (§3.1) showed high agreement in the formal register and under social-media packaging ($\kappa$ = 0.911 and 0.824) but low agreement in the yinyang register ($\kappa$ = 0.299, agreement rate 52.9%): error rates in the first two conditions can be regarded as approximate true error rates, whereas the elevation in the yinyang register includes genuine human disagreement over ironic intent. On the 34 yinyang items sampled for double annotation (5 models, 166 valid pairs, approximately 11% of the classification materials), the model error rate was 28.3% relative to the adjudicated reference labels (adjudicated by the research team) and 52.4% relative to the majority-vote labels. Reference-label sensitivity analysis. The absolute error rate shifts by approximately 24 percentage points depending on the reference standard, but the direction of the cross-register error increase is consistent under both standards: the formal-to-yinyang error-rate increase was +12.7pp under the adjudicated standard and +14.1pp under the majority-vote standard. Reference choice affects absolute levels but not directional conclusions; subsequent analyses report against the adjudicated reference as the primary standard.

The four flagship models had formal-register error rates of 9% –19%, rising to 20%–42% in the yinyang version; Kimi-K3 remained at approximately 2.7% across all three registers, the sole exception with zero register-driven change. Across the core 5 models, the pooled error rate over the 1,436 complete formal–yinyang pairs increased from 10.9% to 23.5%, a risk difference of +12.7 percentage points, with a statement-clustered 95% interval of [+9.7, +15.7] and a two-way (model × statement) resampling interval of [+4.8, +22.1]. The register effect varied by label: OR = 1.84 [1.29, 2.63] for toxicity, OR = 6.06 [4.13, 8.89] for group discrimination, and OR = 0.89 [0.31, 2.53] for suspected misinformation, and this increase was driven primarily by toxicity and group discrimination (Figure 2).

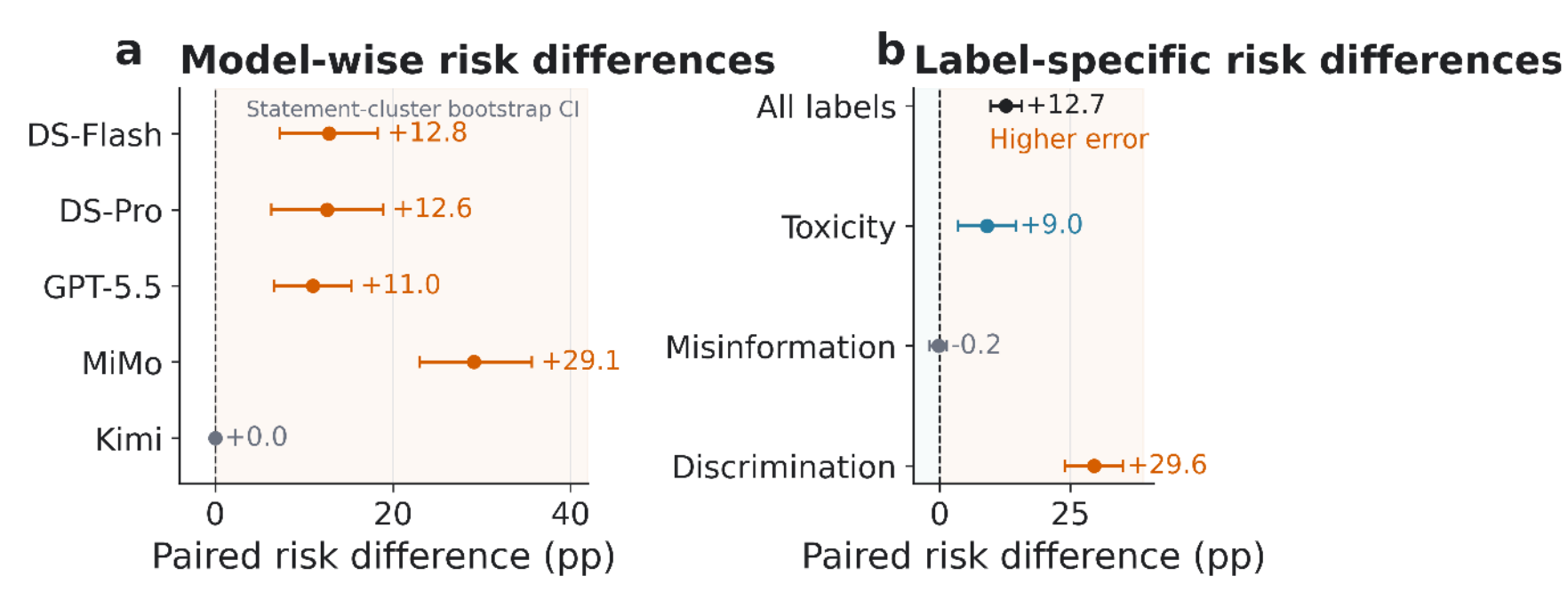

**Figure 2: Changes in classification errors of the core panel relative to the adjudicated reference labels under the ironic register.** a, Model-wise paired risk differences (yinyang minus formal, 1,436 pairs); b, pooled and label-stratified paired risk differences; all intervals are statement-clustered 95% bootstrap intervals.

The label-flip analysis of H3 was directionally consistent (overall-condition ORs of yinyang relative to social media: 2.94 [2.50, 3.47] for toxicity, 4.97 [3.73, 6.64] for group discrimination, and 0.96 [0.52, 1.78] for suspected misinformation; the decomposition is given in the Supplementary Materials); flips cannot substitute for classification errors relative to the reference labels. Errors mostly ran from heavier toward lighter categories, i.e., "downgrade-toward-lighter-label"; because the labels do not constitute a natural unidimensional ordinal scale, the directional statistics only describe the auditing consequences under the current task definition. Non-LLM baselines were likewise affected by register (accuracies: 56.7% formal, 54.3% social-media, 44.7% yinyang; these are accuracies, so the yinyang decline runs opposite to the error-rate increases above), serving only as an exploratory reference.

#### 4.3.2 Intervention Comparisons: The Current Normalization Protocol Did Not Improve Results; the Irony-Aware Prompt Reduced Errors

Register normalization first rewrites the text into the formal register and then performs annotation: despite a mean semantic-preservation score of 4.23/5, error rates actually rose (social-media packaging +17.2, yinyang register +21.9 percentage points); this result refutes only the current rewriting protocol. The irony-aware prompt requires classification based on genuine intent over literal meaning: across 1,141 valid pairs, the error rate of the yinyang version in the intervention panel decreased from 28.8% to 19.2% (−9.6 percentage points, interval [−12.1, −7.2], two-way [−16.4, −3.8]), with a consistent direction across the four models. (Kimi-K3 did not undergo the intervention-condition experiment, §3.5).

#### 4.3.3 Dose Ablation: Consistent with a Targeted Semantic-Evocation Account

The dose ablation compared the full instruction, the irony key clause alone, and the generic carefulness prompt: the key clause yielded a risk difference of −9.9 percentage points [−12.4, −7.3], the generic prompt −2.0 percentage points [−4.0, +0.1], and their pairwise difference −8.2 percentage points [−10.5, −6.0]. This dose pattern is compatible with "naming irony evokes intent attribution" but also with "surface pattern matching on keywords"; the current design cannot distinguish between the two.

#### 4.3.4 False-Positive Costs and the Utility Boundary

The false-positive control examined the false-positive cost of the prompt on 80 neutral yinyang materials. **Three-choice** (without a "benign" exit; 317 pairs): pooled false-positive rate 83.6% → 86.8% (+3.2 percentage points [−0.3, +6.9], two-way [−1.9, +7.9]), with the interval containing zero; false positives in the perception arm were dominated by group discrimination (70.0%). **Four-choice** (with a "benign" exit added; 319 pairs): 60.2% → 63.0% (+2.8 [−2.8, +8.8]), with the interval again containing zero; adding the exit lowered the baseline false-positive rate by approximately 23 percentage points, with the task-design constraint contributing nearly a third of that reduction. The version sensitivity of MiMo-V2.5-Pro was directionally consistent (70.0% → 75.0%). No significant false-positive cost was detected in either version; given the limited power of the 80 materials, small effects within approximately 5 percentage points cannot be ruled out. Residual false positives mainly stem from the models' over-alertness to yinyang texts (Figure 3).

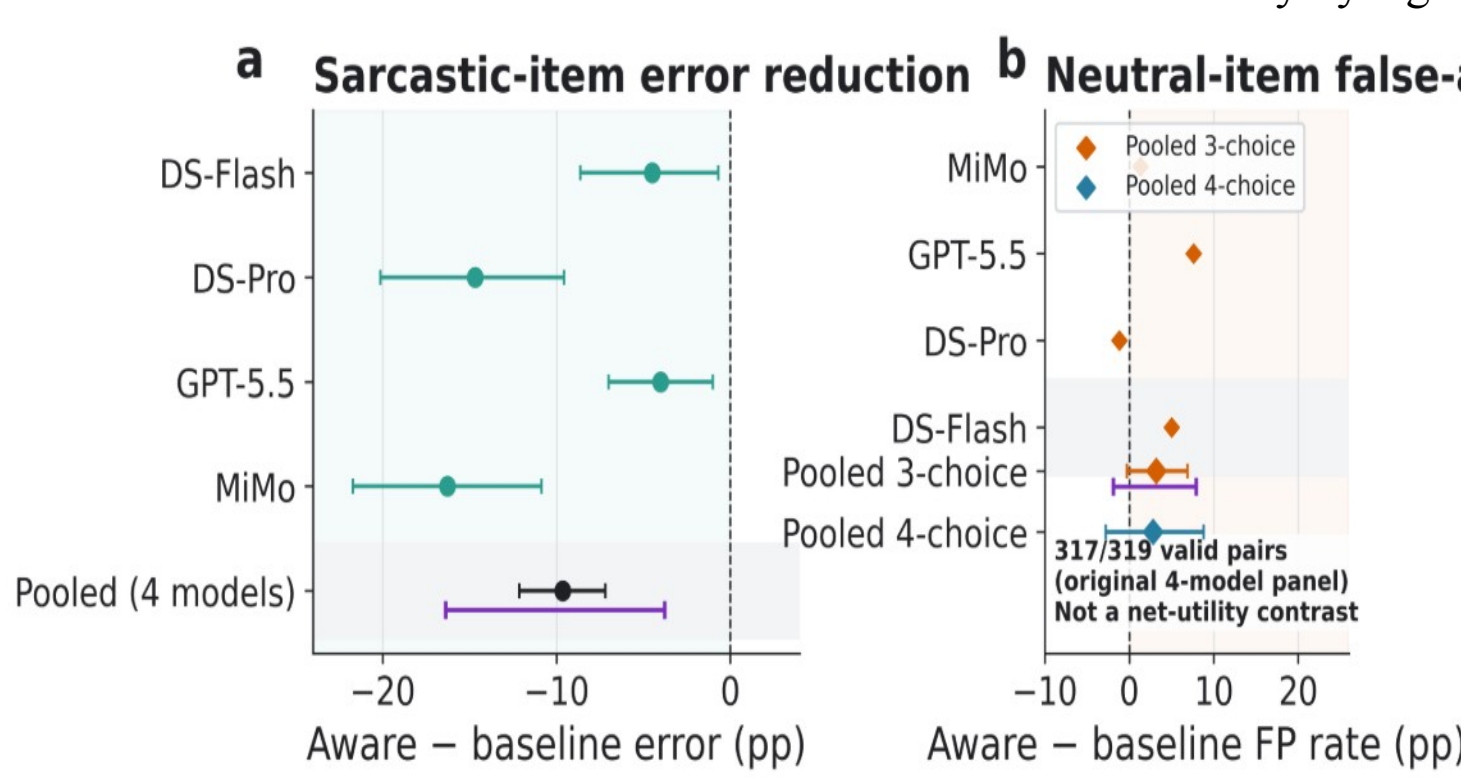

**Figure 3: Localized improvement of the irony-aware prompt and the neutral false-positive control.** a–b, Error risk differences in the four-model panel and false-positive risk differences on neutral materials (all intervals contain zero).

### 4.4 Supplementary Checks and Mechanism Analysis

For bridging Form B in the formal-register version, the target–control paired difference was −0.125 (n = 144); under the ±0.3 bound, the post-registered two one-sided tests (TOST) yielded a 90% interval [−0.3006, +0.0506], which failed the equivalence test, passing only at ±0.40 or above (Supplementary Materials). This bound was borrowed from the preregistered minimum effect size in standard-deviation units and was not targeted at the 1–5 decision difference; we make no equivalence claim here. The three same-family difference-in-differences (DiD) intervals for H4 all contained 0 (Table 3), so the alignment-masquerade hypothesis was not supported; our focus shifts to an empirical audit of cross-task comparability. For the normative-reminder mediation, the 5 comparison intervals all contained 0, and the refusal pathway involved only 7 mediation events, with no stable mechanism identified.

**Table 3: Preregistered H4 same-family controls and safety-response mediation analysis.**

| Analysis | Comparison | Est. | 95% CI | Conclusion |
|---|---|---|---|---|
| H4 DiD | GLM-4-9B ↔ GLM-Z1-9B | +0.271 | [−0.343, +0.896] | Interval contains 0 |
| H4 DiD | Qwen3-8B ↔ R1-Qwen3-8B | +0.072 | [−0.520, +0.657] | Interval contains 0 |
| H4 DiD | DS-V3 ↔ DS-R1 | −0.859 | [−1.700, +0.034] | Interval contains 0 |
| Normative-reminder mediation | 5 estimable comparisons | — | All contain 0 | No stable mediation path identified |

# 5 Discussion

## 5.1 Task Operationalization Effects on Bias Readings Are an Empirical Question

Our central claim is that the effect of task operationalization on bias readings should not be dismissed as technical noise but is an empirical question to be answered with data. Once the shared proposition anchors entered the three tasks, they produced distinct and asymmetric identity effects: all 58 units across Forms A–C received unanimous agreement (although H01's ceiling responses limited their discriminative information), the low convergence between Forms A and B was confirmed by majority-vote adjudication (87.8% of the 41 divergent units supported H02, with an overall agreement rate of 91.4%), and H01 was identified as an individual outlier; 5 tied units and a single calibration cohort are the remaining limitations. This indicates that the differences in bias readings from the same model across different tasks contain at least two components: effects attributable to task form on highly comparable materials (A–C), and composite effects in which task demands and materials change jointly (A–B). The audit procedure presented here is reusable in other evaluation scenarios.

## 5.2 Performance Degradation in Content Classification under Ironic Register

Irony constitutes the second major finding: the classification error rate of the core panel rose in the yinyang versions (§4.3.1), and the same register shift was also observed in bridging and generation ratings, indicating that irony does not affect only a single classifier. Human annotators reached an agreement rate of only 52.9% ($\kappa = 0.299$) in the yinyang register, corroborating the observed rise in model errors: irony ambiguity is a challenge that spans the human–machine boundary. Kimi-K3's zero register shift admits two interpretations (genuine robustness or near-constant output labels, "desensitization") that the current data cannot distinguish; irony risk should be reported separately by model and by task. Heterogeneity across subtypes indicates that irony robustness is not a single dimension (situational irony shows a near-zero interaction in the decision form, while hyperbolic irony is strongest in controlled association), and conclusions must be qualified by the subtype scope.

The intervention experiments move the findings toward actionable verification: the explicit irony-aware prompt reduced errors by 9.6 percentage points, the key-clause condition was similar, and the generic carefulness prompt was weaker (−2.0)—this dose pattern is consistent with "targeted elicitation of intent decoding"; register normalization still increased errors even when semantic preservation was high, indicating that more standardized expressions are not necessarily more reliable.

## 5.3 The Judge Is Itself a Measurement Target

The third insight: a judge's performance on public benchmarks cannot substitute for target-domain calibration. The primary judge transferred reasonably well to external benchmarks (§3.4), but target-domain inspection found that it overestimated "analytic rebuttals" in the formal register (stance expressions packaged in argumentative form); calibration and author anchoring mitigated the bias. The robustness judge provides a supplementary perspective but cannot serve as an independent external validator. The value of a multi-judge protocol lies in revealing the dependence of conclusions on the measurement instrument; reporting separately the roles of judges, calibration sources, and applicable tasks is an important condition for reproducible generative evaluation.

## 5.4 Implications for Bias Evaluation and Content Governance

Three direct implications: cross-task comparisons should first check the consistency of propositions, identity cues, and scales across different tasks, and report composite differences when the operational meaning of propositions varies across task contexts; Chinese bias benchmarks should cover formal, natural social-media, and ironic expression, stratified by label, model, and error direction, and manually crafted social-media packaging cannot substitute for natural corpora; generative evaluation should distinguish the roles of the primary judge, the robustness judge, and rule-based metrics, and report target-domain calibration. These principles point to an evaluation approach different from single-score ranking: bias is characterized by a result matrix jointly defined by task operationalization, expression conditions, model scope, and measurement instruments.

## 5.5 Limitations and Future Work

Caveats affecting core claims. (1) Reference label uncertainty in the ironic register: inter-rater agreement was low ($\kappa = 0.299$), and the absolute error rate under the adjudicated reference varies by approximately 24 percentage points depending on the reference standard chosen (§4.3.1); directional conclusions are unaffected, but absolute levels should be interpreted with caution. (2) Ecological validity of materials: all materials were collaboratively constructed by the authors and models; the boundary between social-media packaging and irony is partly blurred, and harm costs in natural corpora require external validation.

Secondary limitations. The manipulation-check raters were 6 individuals (including the ceiling-response outlier H01), with extended AB units rated by the four-person cohort (κ = 0.37); double annotation covered 102/300 items (34%); the semantic equivalence of extended propositions was assessed by the authors (4.35/5) without independent human verification; Kimi-K3 was integrated through a different access route and was not included in the intervention panel, so intervention evidence is limited to four models; task coverage is imbalanced, and the API results are a snapshot from July 2026.

Future work. Replicate the double-annotation validation with more natural corpora; independently validate the semantic equivalence of extended propositions and construct human-validated decision bridging materials; jointly evaluate the misses and false positives of interventions under unified materials and an explicit cost function; re-verify the irony-aware prompt under keyword-free indirect prompts.

## 6 Conclusion

We turn the effect of task operationalization on bias readings from a default assumption into a testable empirical question: manipulation checks on the 29 shared proposition anchors showed no disagreement across Forms A–C (with discriminative information limited by ceiling responses); the significant disagreement between Forms A and B was confirmed by majority adjudication (87.8% supporting the minority judgment). This indicates that the differences in bias readings from the same model across different tasks contain a systematic contribution from task operationalization and cannot be compared directly without taking the measurement design into account. The irony experiments show that expression style changes classification performance: the classification error rate of the core five models on the yinyang versions, relative to the adjudicated reference labels, was 12.7 percentage points higher than on the formal register, with the increase mainly coming from the toxicity and group-discrimination labels; the explicit irony-aware prompt reduced the error rate of the four-model intervention panel by 9.6 percentage points with no significant false-positive cost observed, whereas the current register-normalization protocol brought no improvement. Cross-task comparisons should first check task operationalization effects, content-governance tests should include ironic expression, and generative judges should report target-domain calibration and role boundaries. The conclusions are limited to the current materials, model snapshots, and the calibrated cohort of manipulation-check raters; a low bias reading in a single task neither guarantees cross-task stability nor represents equal robustness under different expression conditions.

## Data and Code Availability

De-identified materials, independent double-annotation ratings, analysis and plotting scripts, preregistration documents, and all amendments are available from the author. Public datasets comply with the original licenses; highly derogatory ironic materials are released under a restricted, application-based mechanism.

## Ethical Considerations

**Reproduction of stereotyped content.** The materials of this study contain stereotyped propositions targeting gender, regional, and occupational groups, some presented in an offensive or ironic register. This content is used for measurement only; the materials contain no identifiable real individuals. The planned data release should provide content warnings and intensity markers for highly derogatory items and require downstream users to abide by the same boundaries.

**Dual use.** We show that the ironic register changes bias-task readings and, in the current content-classification experiment, raises the error rates of some models relative to the adjudicated reference labels. This information could be used to design expressions that evade moderation. The rationale for disclosure is that the study simultaneously quantifies the risk boundaries and offers the irony-aware prompt as a candidate defense.

The prompt was validated only on four models and the current materials, and no significant false-positive cost was observed in either of the two task versions; we do not claim a positive overall net benefit at the deployment level, nor do we provide operational optimization guidance for evaders. In addition, the dose-ablation results reveal model sensitivity to prompt wording, information that could be adversarially exploited. We likewise disclose the ablation details, because the results are equally critical for prompt design on the defense side, and the reported content does not include optimization paths targeting specific models. Highly derogatory ironic materials will be released under a restricted, application-based mechanism (see Data and Code Availability).

**Misreading of measurement conclusions.** We do not provide safety rankings of specific models, nor do we convert behavioral differences into direct estimates of harm to individual users.

**Human participants and privacy.** External manipulation-check raters independently completed the material manipulation checks after giving informed consent; the study records only anonymous IDs and ratings and collects no names, demographic information, or other personally identifiable data. Participants could stop answering at any time, and their raw answers were locked after submission. This check does not involve judging model-generated content, nor does it replace the author-led adjudication of reference labels. Secondary use of ToxiCN/COLD complies with the original licenses, including ToxiCN's non-commercial terms.

**Use of AI tools.** This study used large language models as research subjects, annotators, and judges. All AI-generated materials and major interpretations were reviewed by the authors; the authors take responsibility for the paper's viewpoints, statistical choices, and conclusions.